\documentclass[10pt,twocolumn,letterpaper]{article}

\usepackage[pagenumbers]{cvpr} % To force page numbers, e.g. for an arXiv version

\usepackage[table,xcdraw]{xcolor}

\definecolor{cvprblue}{rgb}{0.21,0.49,0.74}
\usepackage[pagebackref,breaklinks,colorlinks,citecolor=cvprblue]{hyperref}
\usepackage{multirow}
\usepackage{pifont}
\usepackage[table,xcdraw]{xcolor}
\usepackage{graphicx}

\usepackage{tikz} % 用于绘制圆圈

\definecolor{pltblue}{RGB}{174, 199, 232}
\definecolor{pltorange}{RGB}{255, 229, 204}
\definecolor{pltgreen}{RGB}{204, 229, 204}
\definecolor{pltred}{RGB}{229, 204, 204}
\definecolor{pltpurple}{RGB}{239, 218, 230}

\definecolor{tabblue}{HTML}{1f77b4}
\definecolor{taborange}{HTML}{ff7f0e}
\definecolor{tabgreen}{HTML}{2ca02c}
\definecolor{tabred}{HTML}{d62728}
\definecolor{tabpurple}{HTML}{9467bd}
\definecolor{tabpink}{HTML}{ff0080}

\definecolor{cblue}{RGB}{173, 201, 233}
\definecolor{clblue}{RGB}{222, 234, 246}
\definecolor{corange}{RGB}{255, 152, 67}
\definecolor{lorgange}{RGB}{255, 221, 149}

\def\paperID{213} % *** Enter the Paper ID here
\def\confName{CVPR}
\def\confYear{2025}

\title{SegEarth-OV: Towards Training-Free Open-Vocabulary Segmentation for Remote Sensing Images}

\makeatletter
\def\thanks#1{\protected@xdef\@thanks{\@thanks
        \protect\footnotetext{#1}}}
\makeatother

\author{Kaiyu Li$^1$, Ruixun Liu$^1$, Xiangyong Cao$^{1\dag}$, Xueru Bai$^2$, Feng Zhou$^2$, Deyu Meng$^1$, Zhi Wang$^1$\thanks{\dag Corresponding author.} \\
$^1$Xi’an Jiaotong University \; $^2$Xidian University\\
{\tt\small likyoo.ai@gmail.com, liuruixun6343@gmail.com, caoxiangyong@mail.xjtu.edu.cn} \\
{\tt\small xrbai@xidian.edu.cn, fzhou@mail.xidian.edu.cn, dymeng@mail.xjtu.edu.cn, zhiwang@xjtu.edu.cn} \\
Project: \url{https://likyoo.github.io/SegEarth-OV}
}

\begin{document}
% \maketitle

\twocolumn[{%
\renewcommand\twocolumn[1][]{#1}%
\maketitle
\vspace{-8mm}
\begin{center}
  \centering
  \captionsetup{type=figure}
  \includegraphics[width=\linewidth]{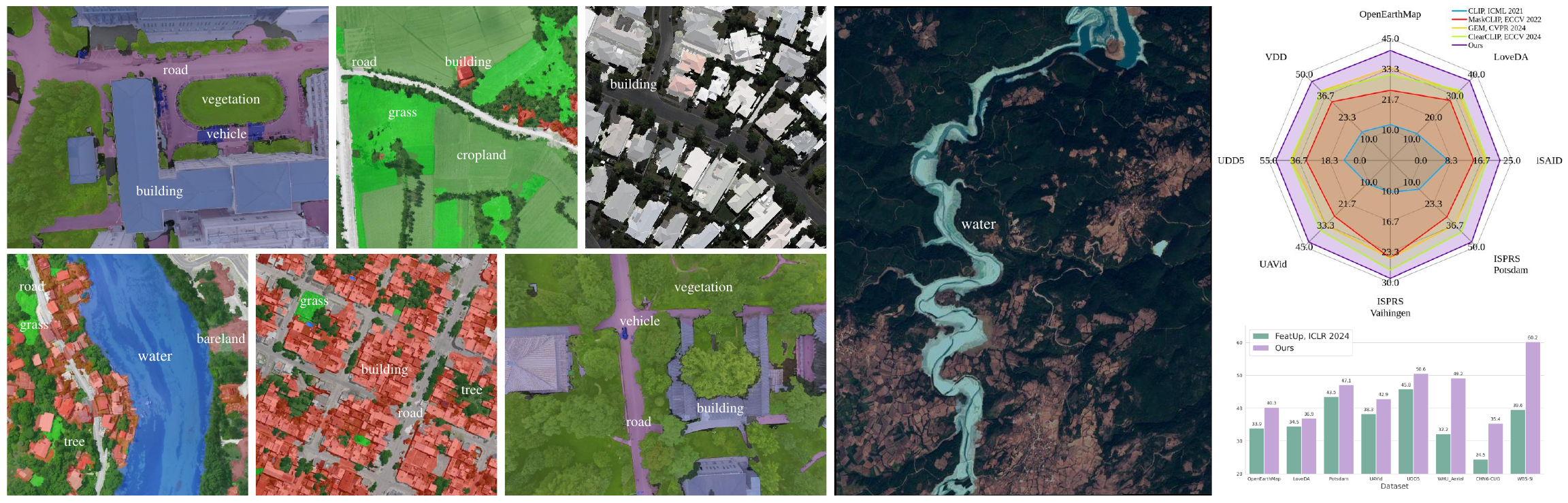}
  \captionof{figure}{Visualization and performance of SegEarth-OV on open-vocabulary semantic segmentation of remote sensing images. We evaluate on 17 remote sensing datasets (including semantic segmentation, building extraction, road extraction, and flood detection tasks), and our SegEarth-OV consistently generates high-quality segmentation masks.}
  \label{fig:fig_title}
\end{center}%
}]

\begin{abstract}
Remote sensing image plays an irreplaceable role in fields such as agriculture, water resources, military, and disaster relief. Pixel-level interpretation is a critical aspect of remote sensing image applications; however, a prevalent limitation remains the need for extensive manual annotation. For this, we try to introduce open-vocabulary semantic segmentation (OVSS) into the remote sensing context. However, due to the sensitivity of remote sensing images to low-resolution features, distorted target shapes and ill-fitting boundaries are exhibited in the prediction mask. To tackle this issue, we propose a simple and general upsampler, SimFeatUp, to restore lost spatial information in deep features in a training-free style. Further, based on the observation of the abnormal response of local patch tokens to \texttt{[CLS]} token in CLIP, we propose to execute a straightforward subtraction operation to alleviate the global bias in patch tokens. Extensive experiments are conducted on 17 remote sensing datasets spanning semantic segmentation, building extraction, road detection, and flood detection tasks. Our method achieves an average of 5.8\%, 8.2\%, 4.0\%, and 15.3\% improvement over state-of-the-art methods on 4 tasks. All codes are released.
\end{abstract}

\section{Introduction}
\label{sec:intro}

Remote sensing image has changed the way humans observe and understand the Earth. It enables us to monitor land cover/use types, respond effectively to natural disasters (e.g., fires, earthquakes, floods), gain insight into food and water resources, etc. Among the 17 Sustainable Development Goals (SDGs)\footnote{\url{https://sdgs.un.org/}} issued by the United Nations, remote sensing image can provide important data support for several goals including ``Zero Hunger'', ``Clean Water and Sanitation'', ``Industry, Innovation and Infrastructure'', ``Climate Action'', ``Life on Land'', etc. \cite{toker2024satsynth} Notably, remote sensing data can be considered as a distinct modality in machine learning. It involves more diverse spatial resolutions (from centimeters to kilometers), temporal dimensions (from hours to decades), and object perspectives (overhead and oriented) than natural images. Therefore, solutions designed for other data modalities (e.g., natural images) may be sub-optimal for remote sensing data \cite{rolf2024mission}.

\begin{figure}[t]
  \centering
%   \fbox{\rule{0pt}{2in} \rule{0.9\linewidth}{0pt}}
   \includegraphics[width=1.0\linewidth]{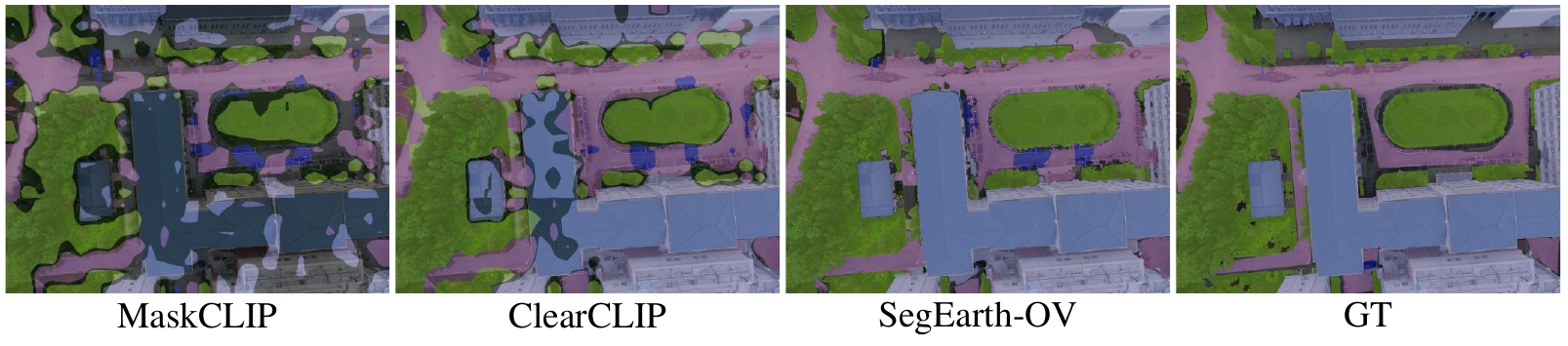}
   \caption{Limitations of state-of-the-art OVSS methods in remote sensing images, the two predictions on the left present distorted target shapes and ill-fitting boundaries. (best viewed digitally with zoom, especially for the edges of the object)}
   \label{fig:motivation}
   \vspace{-4mm}
\end{figure}

In recent years, raw remote sensing images are available from various sources (e.g., QuickBird, WorldView, Landsat, Sentinel), but obtaining large-scale labels is still a challenge due to expensive manual costs. Besides, on the broad surface of the earth, ``stuff'' (e.g., grassland, cropland, roads, forests, etc.) occupies much more area than ``things'' (e.g., buildings, ships, airplanes, etc.) \cite{zanaga2022}. Therefore, for remote sensing images, pixel-level perception, i.e., segmentation, is applied more frequently than instance-level perception, and the demand for pixel-level annotation exacerbates the difficulty of obtaining large-scale labels. OpenStreetMap \cite{haklay2008openstreetmap} is a popular solution that aims to create a freely usable, editable and shareable map of the world. However, the completeness of the annotations in OpenStreetMap is affected by regional income levels, which results in limited data availability \cite{herfort2023spatio}. The rise of vision language model (VLM) brings us new inspirations with its capabilities of open-vocabulary semantic segmentation (OVSS). However, through some exploratory experiments, we find that the solution designed for natural images is sub-optimal on remote sensing images. A notable phenomenon is the presence of distorted target shapes and ill-fitting boundaries in the prediction mask, as shown in \cref{fig:motivation}.

% , as shown in Fig. x.

% On the one hand, through qualitative analysis, we observe the presence of distorted target shapes and ill-fitting boundaries in the prediction masks. On the other hand, through quantitative analysis, we find that compared to natural images, there are more small or even tiny targets in remote sensing images, as shown in Fig. x. 
% \normalsize{\textcircled{\scriptsize{15}}}\normalsize\enspace

% In view of these limitations, we introduce a novel training-free OVSS model (SegEarth-OV), which can also be seamlessly integrated into the other existing method

Empirically, we believe that these issues can be largely attributed to excessively low feature resolution \cite{zheng2023farseg++, guo2022segnext}: In the current CLIP-based OVSS paradigm, the feature maps from CLIP \cite{radford2021learning} are downsampled to $1/16th$ of the original image (ViT-B/16). Hence, in this paper, we propose a simple and general feature upsampler, SimFeatUp, which is trained with the goal of reconstructing content-invariant high-resolution (HR) features on a few unlabeled images, and can upsample arbitrary remote sensing image features after training. Thanks to this property of SimFeatUp, it can be used as a universal external unit for training-free OVSS framework. Further, CLIP is trained at the image level, it uses the \texttt{[CLS]} token as a representation of the entire image, and attaches global properties to the local token \cite{mukhoti2023open, ranasinghe2023perceptual, wang2023sclip}. However, this global property biases local features against patch-level inference in OVSS. We find that a simple subtraction operation of local patch features and global features can effectively reduce global bias. Extensive quantitative and qualitative experiments demonstrate the superior segmentation quality of our method over prior works.

\vspace{2mm}
\noindent\textbf{Contributions:}

\begin{itemize}
    \item We propose SimFeatUp, an general feature upsampler for training-free OVSS, which robustly upsamples low-resolution (LR) features and maintains semantic consistency with image content.
    
    \item We propose an extremely simple and straightforward way to alleviate the global bias problem of CLIP, i.e., executing subtraction operations of local and global tokens.
    
    \item Our final presented model, named SegEarth-OV, achieves state-of-the-art performance on 17 remote sensing datasets spanning semantic segmentation, building extraction, road extraction, and flood detection tasks.
    
\end{itemize}

\section{Related Work}
\label{sec:related_work}

% \begin{figure}[t]
%   \centering
% %   \fbox{\rule{0pt}{2in} \rule{0.9\linewidth}{0pt}}
%    \includegraphics[width=1.0\linewidth]{figures/performance.pdf}
%    \caption{xxx}
%    \label{fig:fig_performance}
% \vspace{-6mm}
% \end{figure}

\noindent\textbf{Vision-Language Model.}
Recently, foundation models, especially visual language models, have energized the field of computer vision. One phenomenal advance is contrastive language-vision pretraining, i.e., CLIP \cite{radford2021learning}, which elegantly bridges the gap between images and natural language. By training with massive data in a multimodal embedding space, CLIP gains strong transfer capabilities, achieving leaps in zero-shot learning and making OV learning possible \cite{wu2024towards}. Subsequently, related research has gradually emerged, from the data \cite{cherti2023reproducible, yang2023alip, xu2023metaclip, singh2024synthetic}, training \cite{li2023scaling, yang2023alip, fan2024improving} or model \cite{li2022blip, li2023blip} side. However, CLIP focuses only on global \texttt{[CLS]} tokens, and even though patch-level tokens can be generated, they are inevitably contaminated by global bias \cite{mukhoti2023open, ranasinghe2023perceptual, wang2023sclip}, which is detrimental to dense prediction. In addition, several remote sensing VLMs emerge, they adapt general VLMs to remote sensing contexts \cite{liu2024remoteclip, zhang2023rs5m, wang2024skyscript, pang2024h2rsvlm} or mine the characteristics of remote sensing data \cite{hong2024spectralgpt, pang2024hsigene}.

% In this work, a extremely simple method is proposed to alleviate the global bias without any training.

% In addition, several remote sensing VLMs emerge, they adapt general VLMs to remote sensing contexts \cite{liu2024remoteclip, zhang2023rs5m, wang2024skyscript, pang2024h2rsvlm} or mine the characteristics of remote sensing data \cite{hong2024spectralgpt, pang2024hsigene}. Although this work involves the intersection of VLMs with the remote sensing field, it is not the focus of this paper and we discuss it in the Appendix \cref{sec:rs_clip}.

\noindent\textbf{Supervised semantic segmentation.}
% # unet -> fenbianlv
Semantic segmentation aims to discriminate images at the pixel level. The prediction head (aka decoder), as an essential component of segmentation models, is able to upsample LR feature maps into HR predictions. Typical prediction heads contain upsampling operators (e.g., bilinear interpolation, JBU \cite{kopf2007joint}) and HR encoder features (as guidance), e.g., UNet \cite{ronneberger2015u}, UperNet \cite{xiao2018unified}, Semantic FPN \cite{kirillov2019panoptic}, MaskFormer \cite{cheng2021per}, etc. Some works \cite{liu2023learning, lu2022sapa, zhou2024refreshed} focus on dynamic, learnable upsampling operators that make this process content-aware. FeatUp \cite{fu2024featup} constructs a model-agnostic upsampling scheme that uses multi-view consistency loss with deep analogies to NeRFs \cite{mildenhall2021nerf}. \textbf{However, it only explores the condition with labels.} Inspired by FeatUp and built on top of it, the SimFeatUp proposed in this work is able to significantly improve OVSS without any labels.

\noindent\textbf{Open-Vocabulary Semantic Segmentation.}
As VLMs have shown remarkable zero-shot inference in image classification, which naturally extends to semantic segmentation. They empower the segmentation pipeline to recognize seen and unseen categories, and users can segment almost any category in an image using prompt vocabulary \cite{wu2024towards, zhu2024survey}. We divide current CLIP-based OVSS methods into two groups: training-required and training-free. The former allows models to be trained on some base classes in a supervised or weakly supervised manner. Typically, some works \cite{ghiasi2022scaling, mukhoti2023open, ranasinghe2023perceptual, luo2023segclip} try to train a localization-aware CLIP which can naturally make dense predictions, while others \cite{li2022language, ding2022decoupling, xu2023side, xu2023san, cho2024cat, liu2024open} select a subset of the CLIP's pre-trained parameters and/or introduce a limited number of trainable parameters into the frozen CLIP, i.e., fine-tuning the CLIP to adapt to dense prediction on base classes. Still, training-free OVSS methods emphasize tapping into CLIP's inherent localization capabilities with limited surgery of features or structures. MaskCLIP \cite{zhou2022extract} pioneers the removal of query and key projections at the attention pooling layer of CLIP's image encoder. Following it, subsequent studies \cite{li2023clip, wang2023sclip, bousselham2024grounding, lan2024clearclip} adequately explore self-self attention (i.e., \textit{q-q}, \textit{k-k} or \textit{v-v} self-attention), and these modifications somewhat mitigate noisy activations and spatial invariant perception of CLIP. Another stream \cite{shao2024explore, lavg, sun2024clip, barsellotti2024training} is the two-stage method, which first generates category-agnostic mask proposals and then classifies the masks. Besides, some other foundation models (e.g. SAM \cite{kirillov2023segment}, Stable Diffusion \cite{rombach2021highresolution}) can be introduced to enhance the localization ability of CLIP, and these explorations also make sense \cite{lan2024proxyclip, barsellotti2024training, wang2023diffusion}.

Different from previous methods, we focus on the inherent characteristics of remote sensing images rather than the general attributes of natural images. The only contemporaneous work is \cite{cao2024open}, but it is training-required, like \cite{xu2023side, cho2024cat}. Our SimFeatUp component, although it needs to be trained on a few images-only data beforehand, this process is independent of the semantic segmentation process and the trained weights can be used for almost any remote sensing data (like the foundation model in other works \cite{lan2024proxyclip, barsellotti2024training}), so our method can still be seen as a training-free method.

% In general, remote sensing images contain a wider range of background and logarithmic scale spans than natural images \cite{rolf2024mission}. This implies the presence of more small targets, as shown in Fig. X, and thus they place higher demands on the model's localization capabilities. 

\begin{figure}
  \centering
  \begin{subfigure}{0.5\textwidth}
    \centering
    \includegraphics[width=1.0\linewidth]{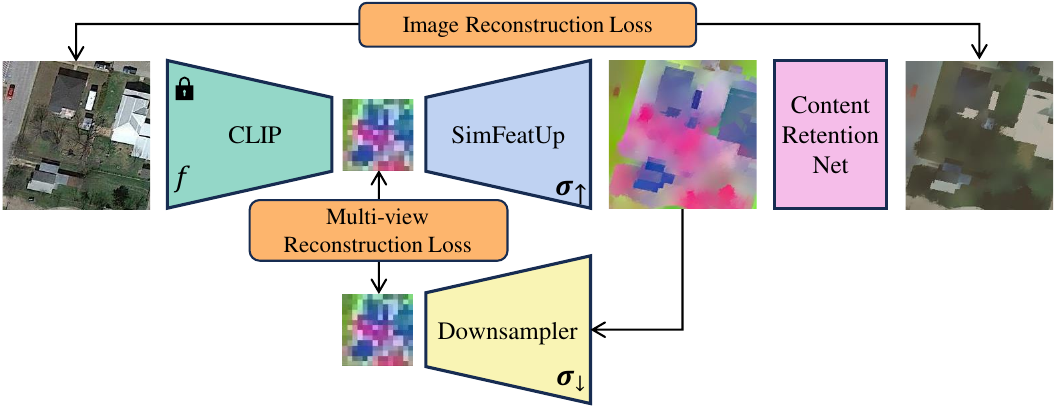}
    \caption{}
    \vspace{2mm}
    \label{fig:sub1}
  \end{subfigure}
  \hfill
  \begin{subfigure}{0.5\textwidth}
    \centering
    \includegraphics[width=1.0\linewidth]{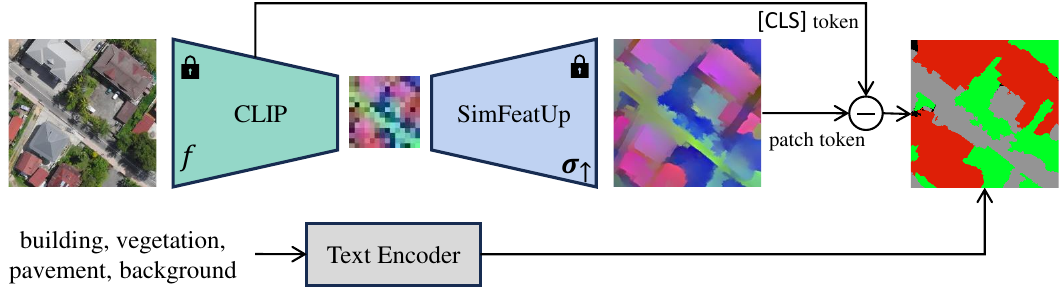}
    \caption{}
    \label{fig:sub2}
  \end{subfigure}
  \caption{Illustration of the proposed method. (a) is the training process of SimFeatUp. CLIP is frozen and only SimFeatUp is useful in reasoning. (b) is the reasoning process of SegEarth-OV. The LR feature maps from CLIP are upsampled by SimFeatUp and then the \texttt{[CLS]} token is subtracted to alleviate global bias. For better presentation, the color renderings follow \cite{fu2024featup}.}
  \label{fig:model}
\vspace{-1em}
\end{figure}

\section{Preliminaries}

\subsection{CLIP}
\label{sec:clip}

In ViT-based CLIP, the image encoder consists of a series of Transformer blocks. Let $X=\left[ x_{cls}, x_1, ..., x_{h\times w} \right] ^{\mathsf{T}} \in \mathbb{R}^{(hw+1, d)}$ denotes the input of the last block, where $h$ and $w$ denote the height and width of the feature map, $d$ denotes the dimention of tokens, and $x_{cls}$ is a learnable global token and the others are local tokens from different image patches. The forward process of this block can be formulated as:

\vspace{-1em}
\begin{equation}
\begin{aligned}
\begin{array}{c}
\boldsymbol{q}=\operatorname{Emb}_{q}(X), \boldsymbol{k}=\operatorname{Emb}_{k}(X), \boldsymbol{v}=\operatorname{Emb}_{v}(X), \\
\boldsymbol{y}=X+\operatorname{SA}\left(\boldsymbol{q}, \boldsymbol{k}, \boldsymbol{v}\right), \\
\boldsymbol{z}=\boldsymbol{y}+\operatorname{FFN}(\operatorname{LN}(\boldsymbol{y})),
\end{array}
\label{eq:attn}
\end{aligned}
\end{equation}

where $\boldsymbol{q}$, $\boldsymbol{k}$, $\boldsymbol{v}$ denote Query, Key, and Value, respectively. $\operatorname{Emb}$ consists of a layer normalization (LN) layer and a linear layer, and FFN denotes the feed-forward network. $\operatorname{SA}$ denotes a standard self-attention module, i.e., $\operatorname{SA}(\boldsymbol{q}, \boldsymbol{k}, \boldsymbol{v}) = \text{softmax}(\frac{\boldsymbol{q} \cdot \boldsymbol{k}^\mathsf{T}}{\sqrt{d}}) \cdot \boldsymbol{v}$. Then, a projection layer maps $\boldsymbol{z}$ to a multi-modal embedding space:

% \vspace{-1em}
\begin{equation}
\begin{aligned}
\mathcal{O} = \operatorname{Proj}(\boldsymbol{z}),
\label{eq:proj}
\end{aligned}
\end{equation}

where $\mathcal{O} =\left[ o_{cls}, o_1, ..., o_{h\times w} \right] ^{\mathsf{T}} \in \mathbb{R}^{(hw+1, c)}$ denotes the output of the image encoder, $c$ denotes token dimension after the projection layer, and $c < d$. During CLIP training, $o_{cls}$ is used for image-level learning; while during OVSS inference, $\mathcal{O}[1:hw+1]$ is used for patch-level prediction.

% a projection layer maps the $\texttt{[CLS]}$ token of $\boldsymbol{z}$, i.e., $\boldsymbol{z}_{cls}$, to a multi-modal embedding space：

% \begin{equation}
% \begin{aligned}
% \boldsymbol{o} = \operatorname{Proj}(\boldsymbol{z}_{cls}).
% \label{eq:proj}
% \end{aligned}
% \end{equation}

% During OVSS inference, 

\subsection{FeatUp}
\label{sec:featup}

FeatUp \cite{fu2024featup} aims to train a model-agnostic upsampler. It executes an upsampling operation on LR features $\mathcal{O}[1:hw+1]$ from a frozen backbone network via a learnable upsampler $\sigma_{\uparrow}$, and then reconstructs the LR features using a learnable downsampler $\sigma_{\downarrow}$. Its critical insights can be briefly summarized by the following loss function:

\vspace{-1em}
\begin{equation}
\begin{aligned}
\mathcal{L}_{rec}=\left\|\mathcal{O}[1:hw+1] - \sigma_{\downarrow}(\sigma_{\uparrow}(\mathcal{O}[1:hw+1])) \right\|_{2}^{2}.
\label{eq:featup_loss}
\end{aligned}
\end{equation}

FeatUp instantiates $\sigma_{\uparrow}$ as stacked parameterized JBU operators. The upsampled HR feature element is estimated by weighting the neighboring elements of the LR feature. For weight generation, JBU considers two factors, the similarity and distance between neighboring elements and the center element in the guidance feature, corresponding to kernel $k_{range}$ and $k_{spatial}$. For brevity, we omit the multi-view consistency constraint in \cite{fu2024featup}.

\section{Method}

In the following, we introduce SegEarth-OV by first describing SimFeatUp's training, design and explaining why it is suitable for OVSS. Second, we discuss the impact of global token on dense prediction and present our method for alleviating global bias.

\subsection{SimFeatUp}
\label{sec:simfeatup}

\begin{figure}[t]
  \centering
%   \fbox{\rule{0pt}{2in} \rule{0.9\linewidth}{0pt}}
   \includegraphics[width=0.9\linewidth]{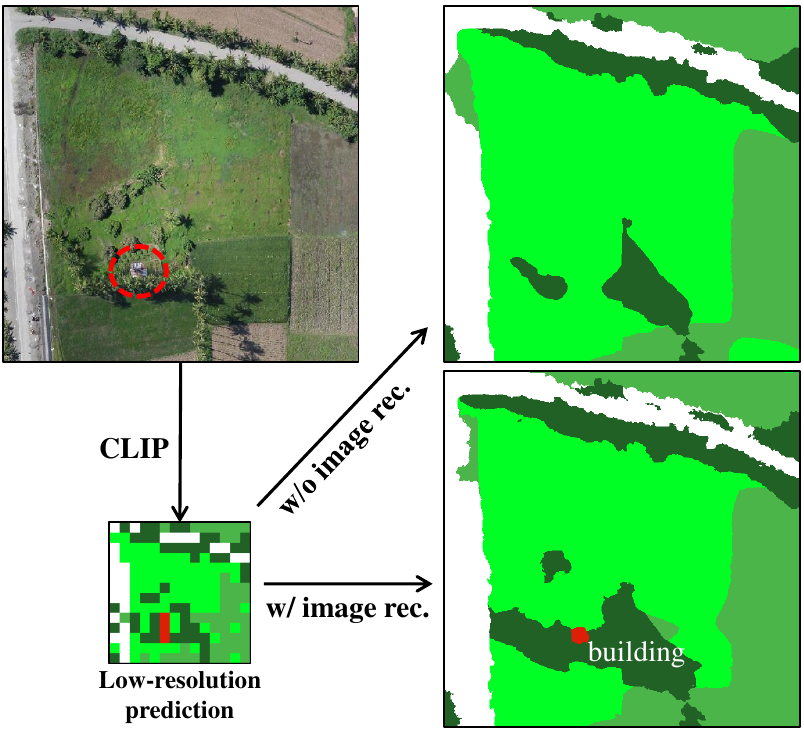}
   \caption{Comparison of with and without image reconstruction loss (\cref{eq:img_rec_loss}). the LR prediction is obtained directly using the output of CLIP (without bilinear interpolation). Color: \textcolor[RGB]{222, 31, 7}{building}, \textcolor[RGB]{34, 97, 38}{tree}, \textcolor[RGB]{75, 181, 73}{cropland}, \textcolor[RGB]{0, 255, 36}{grass}.}
   \label{fig:fig_rec}
\vspace{-1em}
\end{figure}

FeatUp provides us with an excellent training paradigm for general upsamplers. However, it lacks some considerations for the training-free setting, leading it to be sub-optimal for the OVSS task, especially in remote sensing contexts.

\noindent\textbf{Image content retention.} As described in \cref{sec:featup}, the goal of FeatUp is to minimize the original LR features and the LR features after the up-down-sampling (i.e. $\sigma_{\downarrow}(\sigma_{\uparrow}(\mathcal{O}[1:hw+1]))$). Since both $\sigma_{\uparrow}$ and $\sigma_{\downarrow}$ are learnable, with such a weak constraint, the up-down-sampling process becomes a black box, and there is no guarantee that the intermediate HR features are complete and consistent with the original image in content. A direct example is shown in \cref{fig:fig_rec}, where a small building in the original image is present in the LR prediction but disappears in the HR prediction (top right). To solve this issue, we introduce an additional image reconstruction loss to constrain the HR features:

\vspace{-1em}
\begin{equation}
\begin{aligned}
\mathcal{L}_{img}=\left\| I - \operatorname{CRN}(\sigma_{\uparrow}(\mathcal{O}[1:hw+1]))) \right\|_{2}^{2},
\label{eq:img_rec_loss}
\end{aligned}
\end{equation}

where $I$ denotes the input image, $\operatorname{CRN}$ denotes a content retention net. $\operatorname{CRN}$ is a very lightweight network that receives HR features as input and reconstructs the original image. Specifically, $\operatorname{CRN}$ consists of two 2D convolutional layers with activation and a \textit{Tanh} activation layer, where the \textit{Tanh} layer is designed to constrain the output to [-1, 1], cf. VAEs \cite{kingma2013auto}. Finally, the loss for training SimFeatUp consists of $\mathcal{L}_{rec}$ and $\mathcal{L}_{img}$ with a weight $\gamma$, i.e.,

\begin{equation}
\begin{aligned}
\mathcal{L} = \mathcal{L}_{rec} + \gamma \mathcal{L}_{img},
\label{eq:loss}
\end{aligned}
\end{equation}

% where $\gamma$ is the weight of $\mathcal{L}_{img}$.

\noindent
\textbf{Which feature to upsample?} FeatUp takes the final output of CLIP, i.e., $\mathcal{O}[1:hw+1]$ in \cref{eq:proj}, as input to the upsampler. This can work well in training-based settings, e.g., linear probe \cite{alain2016understanding}. However, in training-free OVSS, as described in \cref{sec:related_work}, vanilla self-attention leads to inferior performance. Therefore, the current OVSS method modulates it to self-self attention, and this law also works in remote sensing images. Under this premise, the $\operatorname{SA}$ in \cref{eq:attn} would be replaced by other modules, and direct upsampling of $\mathcal{O}[1:hw+1]$ would lead to the mismatch between training and inference. Motivated by this, we propose to upsample CLIP features at an earlier layer. Specifically, we select the input of the last Transformer block of the CLIP's image encoder, i.e., $X[1:hw+1]$ in \cref{eq:attn}. Further, the high dimension of tokens in $X$ leads to a high cost upsampler. Therefore, we retain the projection layer. Ultimately, the features $\mathcal{O}^{\prime}$ which need to upsample can be formulated as:

\begin{equation}
\begin{aligned}
\mathcal{O}^{\prime} = \operatorname{Proj}(X[1:hw+1]).
\label{eq:which_fea}
\end{aligned}
\end{equation}

\noindent\textbf{Larger upsampling kernel.} We follow the upsampling operator in FeatUp, i.e., the parameterized JBU. As mentioned in \cref{sec:featup}, the upsampling kernels $k_{range}$ and $k_{spatial}$ of the JBU are computed from the elements within a window in the guidance feature. the generation of $k_{range}$ and $k_{spatial}$ can be formulated as follows:

\begin{equation}
\begin{aligned}
k_{spatial}(p, q)=\exp \left(\frac{-\|p-q\|_{2}^{2}}{2 \tau_{spatial}^{2}}\right),
\label{eq:k_spatial}
\end{aligned}
\vspace{-4mm}
\end{equation}

\begin{equation}
\begin{aligned}
&k_{range}(p, q) = \\
&\operatorname{softmax}_{(a, b) \in \Omega}\left(\frac{1}{\tau_{range}^{2}} MLP(G[i, j]) \cdot MLP(G[a, b])\right),
\label{eq:k_range}
\end{aligned}
\end{equation}

where $(p, q)$ denotes the position in the kernel. $\Omega$ denotes a window centered at $(i, j)$ in the guidance feature $G$, which is extracted from HR RGB image. $\tau_{spatial}$ and $\tau_{range}$ are learnable factors. In remote sensing images, unlike natural images, the size of the target presents a logarithmic scale spanning from the meter scale (e.g., trees, gardens) to the kilometer scale (e.g., forests, rangelands) \cite{rolf2024mission}. Therefore, we set larger upsampling kernels to obtain a wider receptive field. Here, we expand the window size to $11 \times 11$, compared to $7 \times 7$ in FeatUp. A possible concern is that a larger receptive field may introduce more irrelevant context, but with $k_{spatial}$, more distant points consistently contribute lower weights, which makes it more reasonable to use larger upsampling kernels.

\noindent\textbf{Simplify.} On the structural side, we simplify the components in FeatUp. In FeatUp, the parameterized JBU modules are stacked 4 times for 16$\times$ upsampling, and the parameters of each JBU module are independent. Although we fed HR features into the $\operatorname{CRN}$ to ensure the integrity of its content, the behavior of each JBU module is indeterminable. Therefore, in SimFeatUp, we change ``JBU\_Stack'' to ``JBU\_One'', i.e., only one parameterized JBU is used for upsampling. If 16$\times$ upsampling is required, then it only needs to repeat the execution 4 times. Further, ``JBU\_One'' significantly reduces the number of trainable parameters in the upsampler and provides the possibility of upsampling arbitrary multiples.

\subsection{Alleviating global bias}
\label{sec:global_bias}

\begin{figure}[t]
  \centering
%   \fbox{\rule{0pt}{2in} \rule{0.9\linewidth}{0pt}}
   \includegraphics[width=1.0\linewidth]{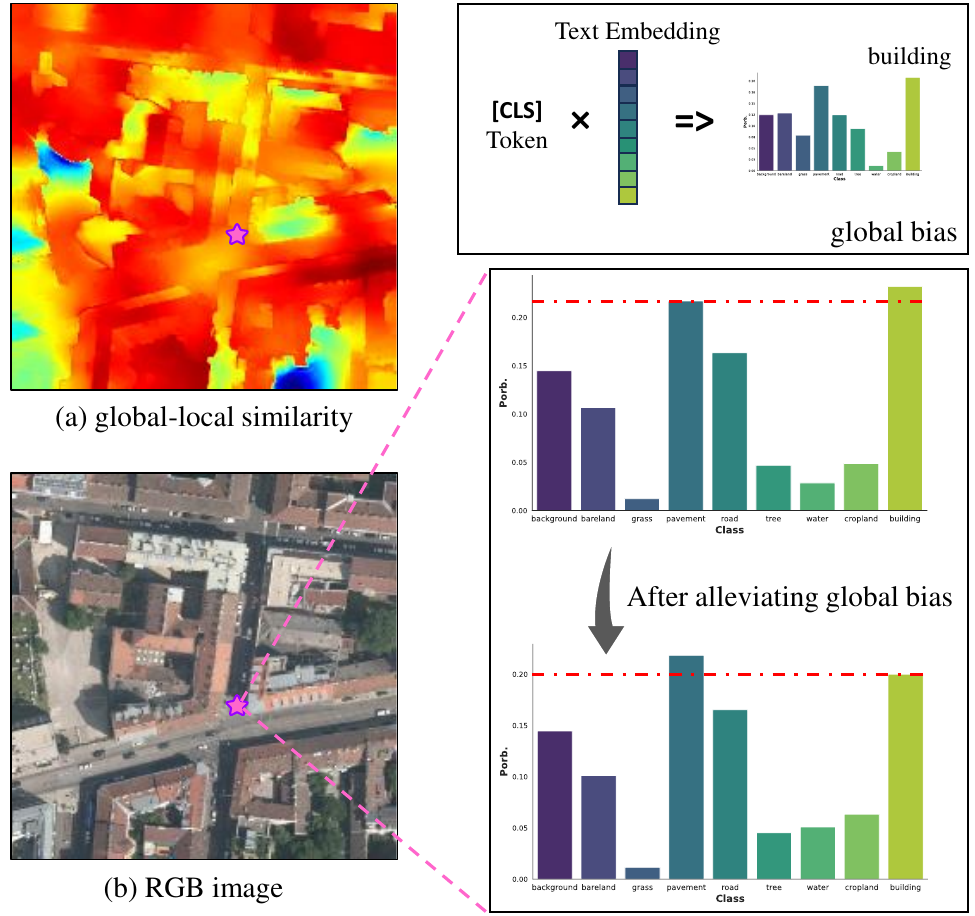}
   \caption{Comparison of before and after alleviating the global bias. (a) is the similarity map of patch tokens and cls tokens, some “non-building” regions also present high response, (b) is the original RGB image. Note that the right-hand histograms stretch the raw values for better presentation.}
   \label{fig:fig_global_bias}
   \vspace{-1em}
\end{figure}

As described in \cref{sec:clip}, in the training phase of CLIP, the \texttt{[CLS]} token, which contains the global information of the whole image, is optimized with the text embedding in the multi-modal space via contrastive learning. However, in the inference phase of OVSS, \texttt{[CLS]} token is generally discarded and only patch tokens are used for similarity computation with the prompt vocabulary. This means that there is a gap between training and inference. Indeed, previous work \cite{mukhoti2023open, ranasinghe2023perceptual, wang2023sclip} also demonstrates that: each local visual token in CLIP focuses on a wide range of positions, and the attention maps typically share similar patterns. This suggests that the global attribute is attached to the patch tokens in CLIP. This property is generally not a concern in classification task, but it significantly impairs performance in dense prediction.

The visualization in \cref{fig:fig_global_bias} demonstrates the above elaboration. We extract the \texttt{[CLS]} token using CLIP for the RGB image in \cref{fig:fig_global_bias}(b), and compute its similarity with the candidate text embeddings. The image is recognized as the building, which is reasonable because the building covers the maximum range in the image. Then, we calculate the similarity of the \texttt{[CLS]} token with patch tokens as shown in \cref{fig:fig_global_bias}(a). The highly responsive regions are not only the regions with buildings, some roads and pavements are also activated, which indicates that the global bias contaminates the local patch tokens. Motivated by this observation, we propose to “subtract” some global bias from the patch token. This solution is very straightforward and simple, it can be formulated as:

\begin{equation}
\begin{aligned}
\hat{\mathcal{O}} = \mathcal{O}[1:hw+1] - \lambda\mathcal{O}[0],
\label{eq:global}
\end{aligned}
\end{equation}

where $\lambda$ denotes a intensity factor. $\mathcal{O}[0]$ is repeated $hw$ times to the same dimension as $\mathcal{O}[1:hw+1]$.

% \subsection{Additional method details}
% \label{sec:additional_method}

% \textbf{Modulated attention.} Previous training-free OVSS methods have extensively explored attention modules at the last layer or layers, from vanilla self-attention to \textit{q-q}, \textit{k-k} or \textit{v-v} self-self attention and their combinations. We find that summation \textit{q-q}, \textit{k-k}, \textit{v-v} as the weights of \textit{v} exhibits a general advantage in remote sensing data. The modulated attention can be formulated as:

% \begin{equation}
% \begin{aligned}
% \operatorname{M-SA}(\boldsymbol{q}, \boldsymbol{k}, \boldsymbol{v}) & = (\text{softmax}(\frac{\boldsymbol{q} \cdot \boldsymbol{q}^\mathsf{T}}{\sqrt{d}}) \\
% &+\text{softmax}(\frac{\boldsymbol{k} \cdot \boldsymbol{k}^\mathsf{T}}{\sqrt{d}}) \\
% &+\text{softmax}(\frac{\boldsymbol{v} \cdot \boldsymbol{v}^\mathsf{T}}{\sqrt{d}})) \cdot \boldsymbol{v}.
% \label{eq:msa}
% \end{aligned}
% \end{equation}

% 两个创新点之间的协同

\section{Experiments}

\begin{table*}
  \caption{Open-vocabulary semantic segmentation quantitative comparison on remote sensing datasets. Evaluation metric: mIoU. \textcolor{tabred}{\textbf{Best}} and \textcolor{tabblue}{\textbf{second best}} performances are highlighted.}
  \label{table_main}
  \centering
  \scalebox{0.9}{
  \begin{tabular}{@{}llcccccccc|c@{}}
    \toprule[1pt]
    Methods & & OpenEarthMap & LoveDA & iSAID & Potsdam & Vaihingen & UAVid$^{img}$ & UDD5 & VDD & Average \\
    \midrule[1pt]
    CLIP \cite{radford2021learning} & {\tiny ICML'21} & 12.0 & 12.4 & 7.5 & 14.5 & 10.3 & 10.9 & 9.5 & 14.2 & 11.4 \\
    MaskCLIP \cite{zhou2022extract} & {\tiny ECCV'22} & 25.1 & 27.8 & 14.5 & 31.7 & 24.7 & 28.6 & 32.4 & 32.9 & 27.2 \\
    % CLIPSurgery&  &  &  &  &\\
    SCLIP \cite{wang2023sclip} & {\tiny arXiv'23} & 29.3 & 30.4 & 16.1 & 36.6 & \textcolor{tabblue}{\textbf{28.4}} & 31.4 & 38.7 & 37.9 & 31.1 \\
    GEM \cite{bousselham2024grounding} & {\tiny CVPR'24} & 33.9 & 31.6 & 17.7 & 36.5 & 24.7 & 33.4 & 41.2 & \textcolor{tabblue}{\textbf{39.5}} & 32.3 \\
    ClearCLIP \cite{lan2024clearclip} & {\tiny ECCV'24} & \textcolor{tabblue}{\textbf{31.0}} & \textcolor{tabblue}{\textbf{32.4}} & \textcolor{tabblue}{\textbf{18.2}} & \textcolor{tabblue}{\textbf{40.9}} & 27.3 & \textcolor{tabblue}{\textbf{36.2}} & \textcolor{tabblue}{\textbf{41.8}} & 39.3 & \textcolor{tabblue}{\textbf{33.4}} \\
    SegEarth-OV & {\tiny Ours} & \textcolor{tabred}{\textbf{40.3}} & \textcolor{tabred}{\textbf{36.9}} & \textcolor{tabred}{\textbf{21.7}} & \textcolor{tabred}{\textbf{47.1}} & \textcolor{tabred}{\textbf{29.1}} & \textcolor{tabred}{\textbf{42.5}} & \textcolor{tabred}{\textbf{50.6}} & \textcolor{tabred}{\textbf{45.3}} & \textcolor{tabred}{\textbf{39.2}} \\
    \bottomrule[1pt]
  \end{tabular}}
  % \vspace{-1mm}
\end{table*}

\subsection{Dataset}

In remote sensing application contexts, not only multi-class semantic segmentation but also extraction of certain land cover types (e.g., buildings, roads, water bodies) is required, e.g., Google's Open Buildings project\footnote{https://sites.research.google/gr/open-buildings/}. Therefore, we select 17 typical datasets covering common semantic segmentation, building extraction, road extraction, and water body segmentation (flood detection) tasks.

\noindent
\textbf{Semantic Segmentation.} We evaluate SegEarth-OV on 8 remote sensing semantic segmentation datasets including OpenEarthMap \cite{xia2023openearthmap}, LoveDA \cite{loveda}, iSAID \cite{waqas2019isaid}, Potsdam, Vaihingen\footnote{https://www.isprs.org/education/benchmarks/UrbanSemLab}, UAVid \cite{LYU2020108}, UDD5 \cite{chen2018large} and VDD \cite{cai2023vdd}. Among them, the first 5 datasets consist of mainly satellite images and the last 3 consist of UAV images. They contain custom foreground classes and a background class. Detailed descriptions of these datasets can be found in Appendix \ref{sec:supp_datasets_ss}.

\noindent
\textbf{Single-class extraction.} We select 4 building extraction datasets (i.e., WHU$^{Aerial}$ \cite{ji2018fully}, WHU$^{Sat.\mathrm{II}}$ \cite{ji2018fully}, Inria \cite{maggiori2017can}, and xBD \cite{gupta2019xbddatasetassessingbuilding}), 4 road extraction datasets (i.e., CHN6-CUG \cite{zhu2021global}, DeepGlobe\footnote{http://deepglobe.org}, Massachusetts \cite{MnihThesis}, and SpaceNet \cite{van2018spacenet}), and 1 flood detection dataset (i.e., WBS-SI\footnote{https://www.kaggle.com/datasets/shirshmall/water-body-segmentation-in-satellite-images}) for the evaluation of single-class extraction. These datasets contain 1 foreground class (building, road or flood) and 1 background class. Detailed descriptions are given in Appendix \ref{sec:supp_datasets_b}-\ref{sec:supp_datasets_f}. 

\noindent
\textbf{Training dataset for SimFeatUp.} SimFeatUp requires only image data for training, moreover, to avoid unfair comparisons, we use a public remote sensing image classification dataset, Million-AID \cite{Long2021DiRS}, which collects images mainly from Google Earth. We randomly selected only 16k of these images to train SimFeatUp.

% It collects images mainly from Google Earth, which contains 1M images across 51 categories.

\subsection{Setup}

\noindent
\textbf{Implementation.} Our implementations are based on MMSegmentation\footnote{https://github.com/open-mmlab/mmsegmentation} toolkit. If not specified, we use the original pretrained weights of CLIP (ViT-B/16) provided by OpenAI. For the text part, we use the \texttt{OpenAI ImageNet template} as input for the text encoder, e.g., ``a photo of a \{class name\}''. In addition, since the definition of certain classes may vary in some datasets, we use slight class rename tricks for all methods. For example, we rename ``clutter'' to ``background'' and ``building'' to \{``building'', ``house'', ``roof''\}, and the highest probability sub-class in \{\} will be the probability of that class. Detailed prompt class names for all datasets are listed in Appendix \cref{table_class_name}. For the image part, we resize input images with a long side of 448 and perform slide inference with a 224 $\times$ 224 window and 112 stride. For SimFeatUp training, we randomly crop 224 $\times$ 224 image patches on the original image. We use two 4090 GPUs to train 1 epoch with batch size set to 8. We retain the multi-view consistency constraint in FeatUp, and random flipping, translation and zoom are applied. For the hyper-parameters mentioned, the value of $\gamma$ is set to 0.1 and $\lambda$ is set to 0.3 for all datasets.

\noindent
\textbf{Evaluation.} We evaluate the semantic segmentation using the mean intersection over union (mIoU) metric. For single-class extraction, the IoU of the foreground class is used.

\noindent
\textbf{Baseline.} We take some lessons from natural image OVSS, which are also suitable for remote sensing scenes: we remove the FFN and residual connection of the last Transformer block, insights from \cite{li2023clip} and \cite{lan2024clearclip}. In addition, the last self-attention is replaced by our modulated attention, i.e., the summation of \textit{q-q}, \textit{k-k} and \textit{v-v} as the weights of \textit{v}:

\vspace{-1em}
\begin{equation}
\begin{aligned}
\operatorname{M-SA}(\boldsymbol{q}, \boldsymbol{k}, \boldsymbol{v}) = \hspace{-1em} \sum_{\boldsymbol{i}\in \{\boldsymbol{q}, \boldsymbol{k}, \boldsymbol{v}\}}\hspace{-1em}\text{softmax}(\frac{\boldsymbol{i} \cdot \boldsymbol{i}^\mathsf{T}}{\sqrt{d}}) \cdot \boldsymbol{v}.
\label{eq:msa}
\end{aligned}
\end{equation}

\subsection{Comparison to State-of-the-art}

Since the proposed SegEarth-OV is a training-free method and there is no previous OVSS method designed for remote sensing images, we select 5 state-of-the-art training-free OVSS models of natural images for comparison, including vanilla CLIP \cite{radford2021learning}, MaskCLIP \cite{zhou2022extract}, SCLIP \cite{wang2023sclip}, GEM \cite{bousselham2024grounding} and ClearCLIP \cite{lan2024clearclip}.

\noindent
\textbf{Semantic segmentation.} As listed in \cref{table_main}, SegEarth-OV achieves the best performance on all 8 semantic segmentation datasets. SegEarth-OV achieves more than 40\% mIoU on 5 datasets and more than 50\% on the UDD5 dataset, which implies that the OVSS method is feasible in remote sensing scenarios. Compared to the previous method, SegEarth-OV achieves a performance gain of more than 5\% on 5 datasets and an average gain of 5.8\% on 8 datasets. On the iSAID dataset, the mIoU of SegEarth-OV is only 21.7\%, which is due to the fine-grained category delineation in this dataset, which covers 16 categories (see Appendix \cref{table_class_name}).

\begin{figure*}[t]
  \centering
%   \fbox{\rule{0pt}{2in} \rule{0.9\linewidth}{0pt}}
   \includegraphics[width=0.9\linewidth]{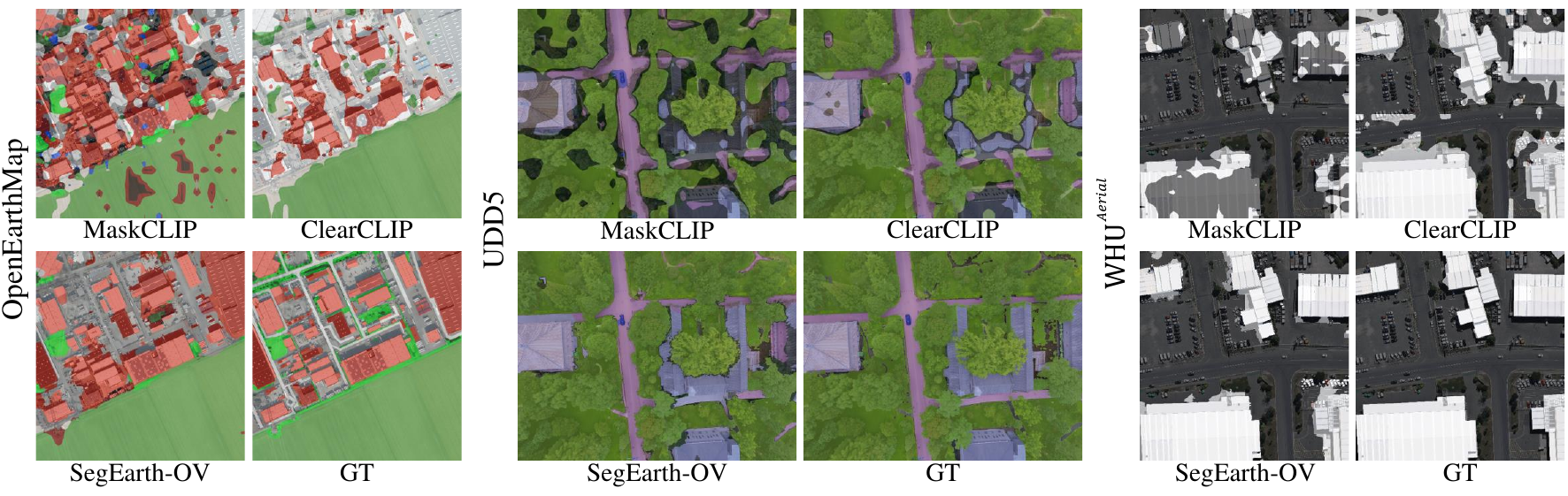}
   \caption{Qualitative comparison between different training-free OVSS methods on OpenEarthMap \cite{xia2023openearthmap}, UDD5 \cite{chen2018large} and WHU$^{Aerial}$ \cite{ji2018fully} datasets. (best viewed digitally with zoom, especially for the edges of the object)}
   \label{fig:samples}
\end{figure*}

\begin{table*}
  \caption{Open-vocabulary building / road / flood extraction quantitative comparison on remote sensing datasets. Evaluation metric: IoU of the foreground class, i.e. building, road or flood. \textcolor{tabred}{\textbf{Best}} and \textcolor{tabblue}{\textbf{second best}} performances are highlighted.}
  \label{table_main_br}
  \centering
  \scalebox{0.85}{
  \begin{tabular}{@{}lcccc|cccc|c@{}}
    \toprule[1pt]
    \multirow{2}{*}{Method} & \multicolumn{4}{c|}{\color{red!50!black}Building Extraction} & \multicolumn{4}{c|}{\color{yellow!50!black}Road Extraction} & \multicolumn{1}{c}{\color{blue!50!black}Flood Detection}\\
    & WHU$^{Aerial}$ & WHU$^{Sat.\mathrm{II}}$  & Inria & xBD$^{pre}$ & CHN6-CUG & DeepGlobe & Massachusetts & SpaceNet & WBS-SI \\
    \midrule[1pt]
    \textbf{\textit{448 $\times$ 448:}} & & & & & & & & & \\
    CLIP \cite{radford2021learning} & 17.7 & 3.5 & 19.6 & 16.0 & 7.7 & 3.9 & 4.9 & 7.1 & 18.6 \\
    MaskCLIP \cite{zhou2022extract} & 29.8 & 14.0 & 33.4 & 29.2 & 28.1 & 13.2 & 10.6 & 20.8 & 39.8 \\
    SCLIP \cite{wang2023sclip} & 33.4 & 21.0 & 34.9 & 25.9 & 21.1 & 7.0 & 7.4 & 14.9 & 32.1 \\
    GEM \cite{bousselham2024grounding} & 24.4 & 13.6 & 28.5 & 20.8 & 13.4 & 4.7 & 5.1 & 11.9 & 39.5 \\
    ClearCLIP \cite{lan2024clearclip} & 36.6 & \textcolor{tabblue}{\textbf{20.8}} & 39.0 & 30.1 & 25.5 & 5.7 & 6.4 & 16.3 & 44.9\\
    SegEarth-OV & \textcolor{tabblue}{\textbf{49.2}} & \textcolor{tabred}{\textbf{28.4}} & \textcolor{tabblue}{\textbf{44.6}} & \textcolor{tabblue}{\textbf{37.0}} & \textcolor{tabred}{\textbf{35.4}} & \textcolor{tabblue}{\textbf{17.8}} & \textcolor{tabblue}{\textbf{11.5}} & \textcolor{tabblue}{\textbf{23.8}} & \textcolor{tabred}{\textbf{60.2}} \\
    \midrule[1pt]
    \textbf{\textit{896 $\times$ 896:}} & & & & & & & & & \\
    SegEarth-OV & \textcolor{tabred}{\textbf{49.9}} & - & \textcolor{tabred}{\textbf{48.9}} & \textcolor{tabred}{\textbf{43.1}} & 32.8 & \textcolor{tabred}{\textbf{20.1}} & \textcolor{tabred}{\textbf{17.2}} & \textcolor{tabred}{\textbf{29.1}} & \textcolor{tabblue}{\textbf{57.9}} \\
    \bottomrule[1pt]
  \end{tabular}}
\end{table*}

\noindent
\textbf{Single-class extraction.} In the building extraction task, the increase delivered by SegEarth-OV is more significant, as listed in \cref{table_main_br}. Considering that the ``building'' class occupies a small area (see Appendix \cref{fig:instance_size}), we evaluate the setup for larger scale images, i.e., resizing the long side of the input image to 896 $\times$ 896. This setup significantly improves the IoU of Inria and xBD, which on the other hand supports our view that spatial detail preservation is essential for remote sensing OVSS. In the road extraction task, although SegEarth-OV achieves the best IoU, overall, the performance of all methods on the 4 datasets is unsatisfactory, with a best IoU of only 35.4\%. There may be two reasons for this phenomenon: (1) The special shape of the road makes it difficult to be extracted in a training-free OVSS manner; (2) The labels of some data are generated based on OpenStreetMap vector shapes with fixed widths attached, which are inherently imprecise. Again, the extraction of roads can generally benefit from larger size images. For the flood detection task, where ``water'' class features can be easily recognized, the IoU of SegEarth-OV is improved by 15.3\% over the previous best method, up to 60.2\%. Due to the small size of the original images in the WBS-SI dataset, resizing to a larger size does not result in a positive gain.

% One observation is that the performance achieved by OVSS is correlated with the spatial resolution, e.g., WHU$^{Aerial}$ and Intra have the highest resolution in the 4 datasets and their IoUs reach 49.2\% and 44.6\%.

\begin{table*}
  \caption{Quantitative comparison of vanilla CLIP and remote sensing CLIPs (ViT-B/32). Evaluation metric: mIoU.}
  \label{table_rsclip}
  \centering
  \scalebox{0.9}{
  \begin{tabular}{@{}llcccccccc|c@{}}
    \toprule[1pt]
    Models & & OpenEarthMap & LoveDA & iSAID & Potsdam & Vaihingen & UAVid$^{img}$ & UDD5 & VDD & Average \\
    \midrule[1pt]
    CLIP \cite{radford2021learning} & {\tiny ICML'21} & 25.7 & 27.2 & 16.2 & 40.0 & \textcolor{tabred}{\textbf{25.1}} & \textcolor{tabblue}{\textbf{31.6}} & \textcolor{tabred}{\textbf{39.7}} & \textcolor{tabblue}{\textbf{39.1}} & 30.6 \\
    \midrule[1pt]
    RemoteCLIP \cite{liu2024remoteclip} & {\tiny TGRS'23} & 18.2 & \textcolor{tabred}{\textbf{37.8}} & \textcolor{tabblue}{\textbf{18.9}} & 21.9 & 22.9 & 16.1 & 27.1 & 28.1 & 23.9 \\
    GeoRSCLIP \cite{zhang2023rs5m} & {\tiny TGRS'24} & \textcolor{tabred}{\textbf{35.0}} & 30.8 & \textcolor{tabred}{\textbf{23.6}} & \textcolor{tabblue}{\textbf{38.0}} & 22.3 & \textcolor{tabred}{\textbf{34.0}} & \textcolor{tabblue}{\textbf{39.1}} & \textcolor{tabred}{\textbf{40.5}} & \textcolor{tabred}{\textbf{32.9}} \\
    SkyCLIP \cite{wang2024skyscript} & {\tiny AAAI'24} & \textcolor{tabblue}{\textbf{28.6}} & \textcolor{tabblue}{\textbf{33.0}} & 15.3 & \textcolor{tabred}{\textbf{41.7}} & \textcolor{tabblue}{\textbf{24.1}} & \textcolor{tabblue}{\textbf{31.6}} & 38.2 & 35.8 & \textcolor{tabblue}{\textbf{31.0}} \\
    \bottomrule[1pt]
  \end{tabular}}
\end{table*}

\noindent
\textbf{Qualitative results.} We present qualitative results for MaskCLIP, ClearCLIP, and SegEarth-OV in \cref{fig:samples}. Some observations are summarized as follows:  (1) There are some incorrect category predictions in MaskCLIP, e.g., \textcolor[RGB]{0, 69, 255}{water} on the road and \textcolor[RGB]{128, 0, 0}{bareland} on the \textcolor[RGB]{75, 181, 73}{cropland}. (2) ClearCLIP can generate correct category predictions, but lacks precise localization capability, with distorted target shapes and ill-fitting boundaries of the prediction mask. (3) SegEarth-OV is capable of generating more fine-grained masks that fit the target edges and maintain correct category discrimination. More visualizations can be found in Appendix \cref{fig:OEM}-\ref{fig:WHU}.

\subsection{Ablation Study and Analysis}

\noindent
\textbf{Plug and Play.} Two key insights of this work, SimFeatUp and global bias alleviation, which can be attached to other OVSS methods as plug-and-play modules. As listed in \cref{table_compare}, a revealing observation is that on both the OpenEarthMap and WHU$^{Aerial}$ datasets, as the base capability of the model improves (from MaskCLIP to ClearCLIP), the increases delivered by our method also improve (\textbf{{\color{green!50!black}$\uparrow$\scriptsize{3.3}}}, \textbf{{\color{green!50!black}$\uparrow$\scriptsize{5.1}}}, \textbf{{\color{green!50!black}$\uparrow$\scriptsize{8.1}}} on OpenEarthMap, \textbf{{\color{green!50!black}$\uparrow$\scriptsize{5.6}}}, \textbf{{\color{green!50!black}$\uparrow$\scriptsize{6.1}}}, \textbf{{\color{green!50!black}$\uparrow$\scriptsize{14.5}}} on WHU$^{Aerial}$). This suggests that our method has the potential to improve localization and discrimination for stronger models. 

% In addition, ``ClearCLIP + ours'' outperforms SegEarth-OV on WHU$^{Aerial}$ and WBS-SI, which suggests that the modulated attention we use is not optimal in some cases, and exploring how to design a better self-attention in training-free OVSS is meaningful. 

% In addition, SimFeatUp, as a plug-and-play module with only $<$ 0.3M parameters, is also effective on natural images, as listed in Appendix \cref{table_natural}.

\begin{table}
  \caption{The proposed method results as a plug-and-play module. ``+ ours'' indicates using SimFeatUp (\cref{sec:simfeatup}) and alleviating the global bias (\cref{sec:global_bias}).}
  \label{table_compare}
  \centering
  \scalebox{0.9}{
  \begin{tabular}{@{}lccc@{}}
    \toprule[1pt]
    Methods & OpenEarthMap & {\color{red!50!black}WHU$^{Aerial}$} & {\color{blue!50!black}WBS-SI} \\
    \midrule[1pt]
    MaskCLIP & 25.1 & 29.8 & 39.8 \\
    \rowcolor{gray!20}
    + ours & 28.4\textbf{{\color{green!50!black}$\uparrow$\scriptsize{3.3}}} & 35.4\textbf{{\color{green!50!black}$\uparrow$\scriptsize{5.6}}} & 48.8\textbf{{\color{green!50!black}$\uparrow$\scriptsize{9.0}}} \\
    SCLIP & 29.3 & 33.4 & 32.1 \\
    \rowcolor{gray!20}
    + ours & 34.4\textbf{{\color{green!50!black}$\uparrow$\scriptsize{5.1}}} & 39.5\textbf{{\color{green!50!black}$\uparrow$\scriptsize{6.1}}} & 53.4\textbf{{\color{green!50!black}$\uparrow$\scriptsize{21.3}}} \\
    ClearCLIP & 31.0 & 36.6 & 44.9 \\
    \rowcolor{gray!20}
    + ours & 39.1\textbf{{\color{green!50!black}$\uparrow$\scriptsize{8.1}}} & 51.1\textbf{{\color{green!50!black}$\uparrow$\scriptsize{14.5}}} & 60.4\textbf{{\color{green!50!black}$\uparrow$\scriptsize{15.5}}} \\
    \bottomrule[1pt]
  \end{tabular}}
\end{table}

\begin{table}
  \caption{Detailed ablation results for each component. ``X''$\uparrow$ indicates upsampling earlier stage features, i.e. \cref{eq:which_fea}. ``+ RS Data'' indicates using  Million-AID \cite{Long2021DiRS} to train the upsampler, before using the images in COCO-Stuff \cite{caesar2018coco}.}
  \label{table_ablation}
  \centering
  \scalebox{0.85}{
  \begin{tabular}{@{}l|c|c|c@{}}
    \toprule[1pt]
    & OpenEarthMap & {\color{red!50!black}WHU$^{Sat.\mathrm{II}}$} & {\color{blue!50!black}WBS-SI} \\
    \midrule[1pt]
    \textit{Baseline} & 32.4 & 22.7 & 46.9 \\
    FeatUp (CLIP) & 33.9 & 20.2 & 39.6 \\
    FeatUp (MaskCLIP) & 33.8 & 25.2 & 45.9 \\
    ``X''$\uparrow$ & 34.6 & 26.0 & 54.2 \\
    + RS Data & 36.0 \textbf{{\color{green!50!black}$\uparrow$\scriptsize{1.4}}} & 26.2 & 56.4 \\
    + JBU\_One & 36.3 \textbf{{\color{green!50!black}$\uparrow$\scriptsize{0.3}}} & 26.0 & 57.1 \\
    + Rec. Image& 37.6 \textbf{{\color{green!50!black}$\uparrow$\scriptsize{1.3}}} & 26.4 & 58.7 \\
    + Alleviate Global Bias & 39.3 \textbf{{\color{green!50!black}$\uparrow$\scriptsize{1.7}}} &  27.9 & 59.5 \\
    + Large Kernel & 40.3 \textbf{{\color{green!50!black}$\uparrow$\scriptsize{1.0}}} & 28.4 & 60.2 \\
    \bottomrule[1pt]
  \end{tabular}}
\end{table}

\noindent
\textbf{Ablation Study.} To assess each of the proposed components, we perform a detailed ablation analysis, as listed in \cref{table_ablation}. FeatUp (CLIP) denotes the original FeatUp upsampler, which provides a 1.5\% improvement on OpenEarthMap but decreases the performance on WHU$^{Sat.\mathrm{II}}$ and WBS-SI (more comparisons between FeatUp and the proposed method are shown in the bottom-right of \cref{fig:fig_title}). FeatUp (MaskCLIP) denotes using $\boldsymbol{v}$ in self-attention as the upsampled feature, which somewhat mitigates the possible negative effects of FeatUp (CLIP). In SimFeatUp, the input feature $X$ of the last block is used to upsample, which presents a significant improvement in all 3 datasets. A substantial improvement is also delivered after replacing the training material for the upsampler from natural images to remote sensing images. ``JBU\_One'' reduces the parameters by nearly $4\times$ while delivering a slight IoU gain (only $<$ 0.3M parameters). The introduction of $\operatorname{CRN}$ with image reconstruction loss brings 1.7\%, 0.4\%, and 1.6\% improvement on 3 datasets, respectively. Note that the $\operatorname{CRN}$ only participates in SimFeatUp's training and is discarded during inference. Global bias alleviation shows significant improvement for all 3 datasets, with an average 1.3\% improvement. Finally, expanding the upsampling kernel to $11\times 11$ also exhibits consistent improvement across all datasets.

\begin{table}
  \caption{OVSS quantitative comparison on natural image datasets. The basic results are cited from \cite{lan2024clearclip}.}
  \label{table_natural}
  \centering
  \scalebox{0.8}{
  \begin{tabular}{@{}lccc|c@{}}
    \toprule[1pt]
    Methods & Context59 \cite{mottaghi2014role} & Stuff \cite{caesar2018coco} & Cityscapes \cite{cordts2016cityscapes} & Average \\
    \midrule[1pt]
    TCL \cite{cha2023learning} & 30.3 & 19.6 & 23.1 & 24.3 \\
    Reco \cite{shin2022reco} & 22.3 & 14.8 & 21.1 & 19.4 \\
    \midrule[1pt]
    MaskCLIP & 26.4 & 16.4 & 12.6 & 18.5 \\
    \rowcolor{gray!20}
    + SimFeatUp & 28.7 & 18.0 & 25.8 & 24.2 \textbf{{\color{green!50!black}$\uparrow$\scriptsize{5.7}}} \\
    SCLIP & 33.0 & 21.1 & 29.1 & 27.7 \\
    \rowcolor{gray!20}
    + SimFeatUp & 34.1 & 22.0 & 30.5 & 28.9 \textbf{{\color{green!50!black}$\uparrow$\scriptsize{1.2}}}\\
    ClearCLIP & 35.9 & 23.9 & 30.0 & 29.9 \\
    \rowcolor{gray!20}
    + SimFeatUp & 37.5 & 25.1 & 30.7 & 31.1 \textbf{{\color{green!50!black}$\uparrow$\scriptsize{1.2}}}\\
    \bottomrule[1pt]
  \end{tabular}}
\end{table}

\noindent
\textbf{Results on natural images.} We evaluate SimFeatUp as an external unit on 3 natural image datasets: PASCAL Context59 \cite{mottaghi2014role}, COCOStuff \cite{caesar2018coco} and Cityscapes \cite{cordts2016cityscapes}. As listed in \cref{table_natural}, after upsampling the visual features of MaskCLIP, SCLIP, and ClearCLIP using SimFeatUp, their mIoUs are improved by 5.7\%, 1.2\%, and 1.2\%, respectively. This reveals the potential of our method to inspire general vision.

\noindent
\textbf{Remote sensing CLIPs for OVSS.} We evaluate the performance of remote sensing CLIPs on OVSS, including RemoteCLIP \cite{liu2024remoteclip}, GeoRSCLIP \cite{zhang2023rs5m}, and SkyCLIP \cite{wang2024skyscript}, which are trained on 0.8M, 5M, and 2.6M remote sensing data, respectively, without changing the model structure of CLIP. Since these works do not provide the ViT-B/16, we uniformly use ViT-B/32. Hence, we repeat the JBU operation 5 times in SimFeatUp. For fair comparison, we train the respective SimFeatUp for each model. As listed in \cref{table_rsclip}, RemoteCLIP performs suboptimally to vanilla CLIP, which indicates that a small amount of domain data diminishes the model's transfer capability. GeoRSCLIP achieves the best performance against SkyCLIP, which suggests that domain VLMs can benefit from more diverse domain-specific data. Moreover, the OVSS task effectively reflects the model's discrimination and localization capabilities, and can serve as an evaluation metric for remote sensing VLMs.

% CLIP Surgery % 全局特征的权重

% 后处理

\section{Conclusion}

In this paper, we present SegEarth-OV, a training-free OVSS method for remote sensing images. The design of SegEarth-OV was motivated by the observation that OVSS methods currently used for natural images do not perform well on remote sensing images. The two key insights of SegEarth-OV, i.e., SimFeatUp and global bias alleviation, exhibit consistent improvements on 17 remote sensing datasets spanning the tasks of semantic segmentation, building extraction, road extraction, and flood detection, well beyond the previous state-of-the-art methods. More importantly, as the first exploration of training-free OVSS method in remote sensing scenarios, this work demonstrates that the OVSS solution is feasible in earth perception tasks even if the VLMs are pre-trained on natural images. We expect that this work will inspire more OVSS methods and more capable remote sensing VLMs, and open up new possibilities for the Earth vision community.
{
    \small
    \bibliographystyle{ieeenat_fullname}
    \bibliography{main}
}

\clearpage
\setcounter{page}{1}
\maketitlesupplementary

\begin{table*}
  \caption{The prompt class name of the evaluation datasets. \{\} indicates multiple prompt vocabularies for one class.}
  \label{table_class_name}
  \centering
  \scalebox{0.9}{
  \begin{tabular}{@{}p{3.7cm}>{\centering\arraybackslash}m{15cm}@{}}
    \toprule[1pt]
    Dataset & Class Name \\
    \midrule[1pt]
    OpenEarthMap & background, \{bareland, barren\}, 
grass, pavement, road, \{tree, forest\}, \{water, river\}, cropland, \{building, roof, house\} \\
    LoveDA & background, \{building, roof, house\}, road, water, barren, forest, agricultural \\
    iSAID &  background, ship, store tank, baseball diamond, tennis court, basketball court, ground track field, bridge, large vehicle, small vehicle, helicopter, swimming pool, roundabout, soccer ball field, plane, harbor \\
    Potsdam, Vaihingen &  \{road, parking lot\}, building, low vegetation, tree, car, \{clutter, background\} \\
    UAVid & background, building, road, car, tree, vegetation, human \\
    UDD5 & vegetation, building, road, vehicle, background \\
    VDD & background, facade, road, vegetation, vehicle, roof, water \\
    WHU$^{Aerial}$, WHU$^{Sat.\mathrm{II}}$, Inria, xBD & background, building \\
    CHN6-CUG, DeepGlobe, Massachusetts, SpaceNet & background, road \\
    WBS-SI & background, water \\
    \bottomrule[1pt]
  \end{tabular}}
\end{table*}

\section{Datasets}
\label{sec:supp_datasets}

\subsection{Semantic Segmentation}
\label{sec:supp_datasets_ss}

\noindent
- \textbf{OpenEarthMap} \cite{xia2023openearthmap} includes worldwide satellite and aerial images with a spatial resolution of 0.25-0.5m. It contains 8 foreground classes and one background class. We use its validation set (excluding xBD data) for evaluation.

\noindent 
- \textbf{LoveDA} \cite{loveda} is constructed using 0.3m images obtained from the Google Earth platform. It contains both urban and rural areas. It contains 6 foreground classes and one background class. We use its validation set for evaluation.

\noindent 
- \textbf{iSAID} \cite{waqas2019isaid} is manily collected from the Google Earth, some are taken by satellite JL-1, the others are taken by satellite GF-2. Its image data is the same as the DOTA-v1.0 dataset \cite{Xia_2018_CVPR}. It contains 15 foreground classes and one background class. We use its validation set for evaluation, which is cropped to 11,644 images by default (patch\_size=896, overlap\_area=384).

\noindent 
- \textbf{Potsdam\footnote{http://www2.isprs.org/commissions/comm3/wg4/2d-sem-label-potsdam.html} and Vaihingen\footnote{http://www2.isprs.org/commissions/comm3/wg4/2d-sem-label-vaihingen.html}}  are for urban semantic segmentation used in the 2D Semantic Labeling Contest. Their spatial resolutions are 5cm and 9cm, respectively, and they contain 5 foreground classes and one background class. We use the validation set for evaluation according to MMSeg's\footnote{https://github.com/open-mmlab/mmsegmentation} setting.

% \subsubsection{Unmanned aerial vehicle data}
\noindent 
- \textbf{UAVid} \cite{LYU2020108} consists of 30 video sequences capturing 4K HR images in slanted views. We treat them as images without considering the relationship between frames, and the classes “static car” and “moving car” are converted to “car”. Therefore, it contains 5 foreground classes and one background class. We use its test set for evaluation, which is cropped to 1020 images (patch\_height=1280, patch\_width=1080, no overlap).

\noindent 
- \textbf{UDD5} \cite{chen2018large} is collected by a professional-grade UAV (DJI-Phantom 4) at altitudes between 60 and 100m. It contains 4 foreground classes and one background class. We use its validation set for evaluation.

\noindent 
- \textbf{VDD} \cite{cai2023vdd} is collected by DJI MAVIC AIR II, including 400 RGB images with 4000*3000 pixel size. All the images are taken at altitudes ranging from 50m to 120m. It contains 6 foreground classes and one background class. We use its test set for evaluation.

\subsection{Building extraction}
\label{sec:supp_datasets_b}

\noindent 
- \textbf{WHU$^{Aerial}$} \cite{ji2018fully} consists of more than 220k independent buildings extracted from aerial images with 0.075m spatial resolution and 450 $km^2$ covering in Christchurch, New Zealand. We use its validation set for evaluation.

\noindent 
- \textbf{WHU$^{Sat.\mathrm{II}}$} \cite{ji2018fully} consists of 6 neighboring satellite images covering 860 $km^2$ on East Asia with 0.45m ground resolution. We use its test set (3726 tiles with 8358 buildings) for evaluation. The original images are cropped to 1000 $\times$ 1000 without overlap.

\noindent 
- \textbf{Inria} \cite{maggiori2017can} covers dissimilar urban settlements, ranging from densely populated areas (e.g., San Francisco’s financial district) to alpine towns (e.g,. Lienz in Austrian Tyrol). It covers 810 $km^2$ with a spatial resolution of 0.3m. We use its test set for evaluation.

\noindent 
- \textbf{xBD} \cite{gupta2019xbddatasetassessingbuilding} covers a diverse set of disasters and geographical locations with over 800k building annotations across over 45k $km^2$ of imagery. Its spatial resolution is 0.8m. We use the pre-disaster satellite data of test set for evaluation.

\subsection{Road extraction}
\label{sec:supp_datasets_r}

\noindent 
- \textbf{CHN6-CUG} \cite{zhu2021global} is a large-scale satellite image data set of representative cities in China, collected from Google Earth. It contains 4511 labeled images of 512 $\times$ 512 size with a spatial resolution of 0.5m. We use its test set for evaluation.

\noindent 
- \textbf{DeepGlobe\footnote{http://deepglobe.org}} covers images captured over Thailand, Indonesia, and India. Its available data cover 362 $km^2$ with a spatial resolution of 5m. The roads are precisely annotated with varying road widths. We use the validation set for evaluation according to the setting in \cite{singh2018self}.

\noindent 
- \textbf{Massachusetts} \cite{MnihThesis} covers a wide variety of urban, suburban, and rural regions and covers an area of over 2,600 $km^2$ with a spatial resolution of 1m. Its labels are generated by rasterizing road centerlines obtained from the OpenStreetMap project, and it uses a line thickness of 7 pixels. We use its test set for evaluation.

\noindent 
- \textbf{SpaceNet} \cite{van2018spacenet} contains 422 $km^2$ of very high-resolution imagery with a spatial resolution of 0.3m. It covers Las Vegas, Paris, Shanghai, Khartoum and is designed for the SpaceNet\footnote{https://spacenet.ai/challenges/} challenge. We use the test set for evaluation according to the setting in \cite{lu2024global}.

\subsection{Flood Detection}
\label{sec:supp_datasets_f}

\noindent 
- \textbf{WBS-SI\footnote{https://www.kaggle.com/datasets/shirshmall/water-body-segmentation-in-satellite-images}} is a satellite image dataset for water body segmentation. It contains 2495 images and we randomly divided 20\% of the data as a test set for evaluation.

% \section{Prompts}
% \label{sec:prompts}

\begin{figure}[t]
  \centering
%   \fbox{\rule{0pt}{2in} \rule{0.9\linewidth}{0pt}}
   \includegraphics[width=0.8\linewidth]{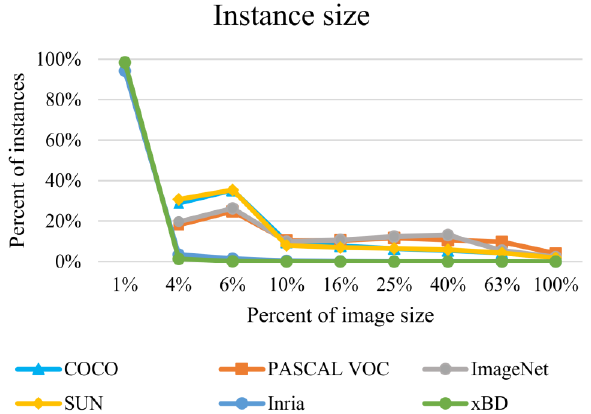}
   \caption{The distribution of instance sizes for natural image datasets (MS COCO, ImageNet Detection, PASCAL VOC and SUN) and remote sensing datasets (Inria and xBD). The data for the natural image is borrowed from \cite{lin2014microsoft}, and the data for Inria and xBD are calculated with the image size at 1024 $\times$ 1024.}
   \label{fig:instance_size}
\end{figure}

\begin{figure*}[t]
  \centering
%   \fbox{\rule{0pt}{2in} \rule{0.9\linewidth}{0pt}}
   \includegraphics[width=0.8\linewidth]{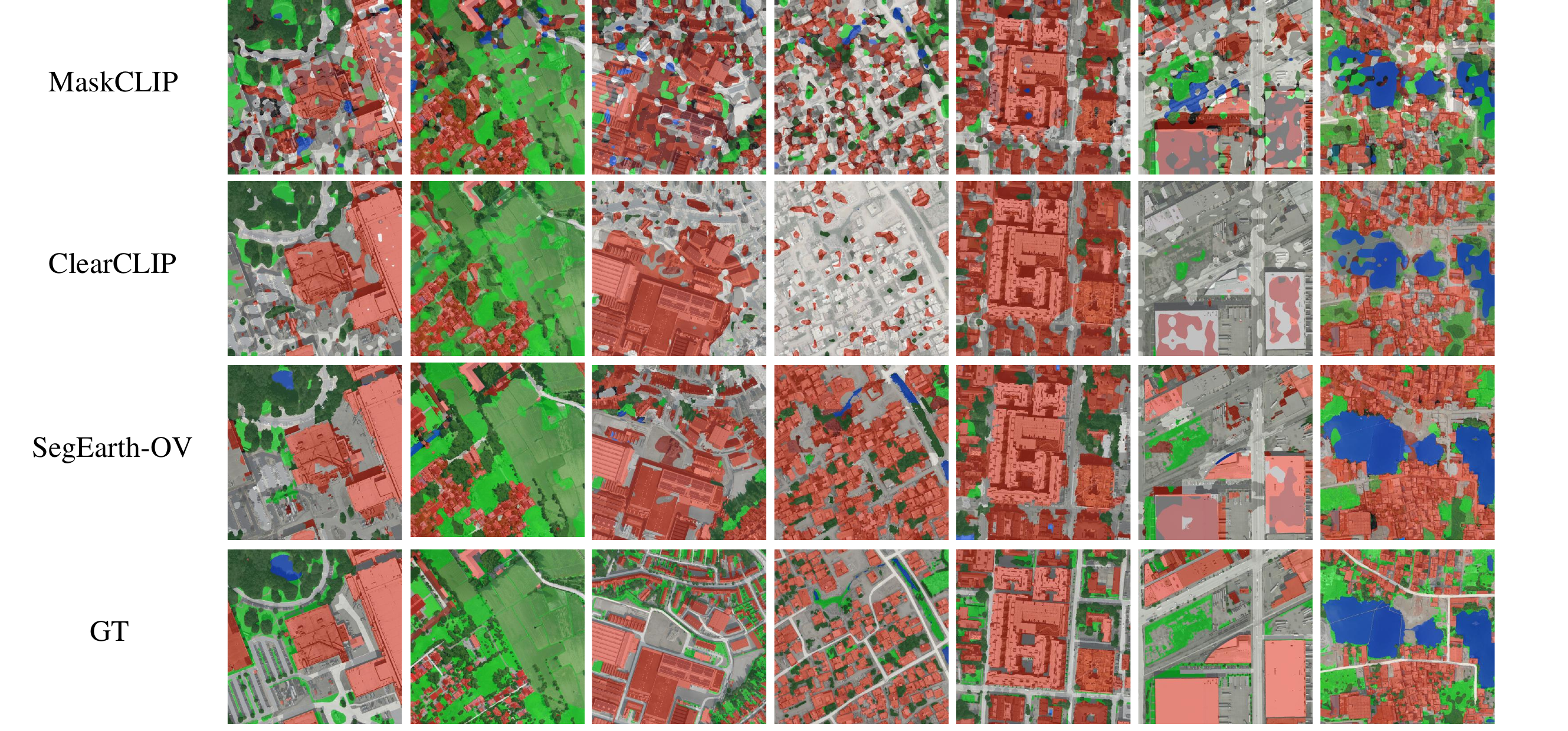}
   \caption{Qualitative comparison between different training-free OVSS methods on OpenEarthMap.}
   \label{fig:OEM}
\end{figure*}

\begin{figure*}[t]
  \centering
%   \fbox{\rule{0pt}{2in} \rule{0.9\linewidth}{0pt}}
   \includegraphics[width=0.8\linewidth]{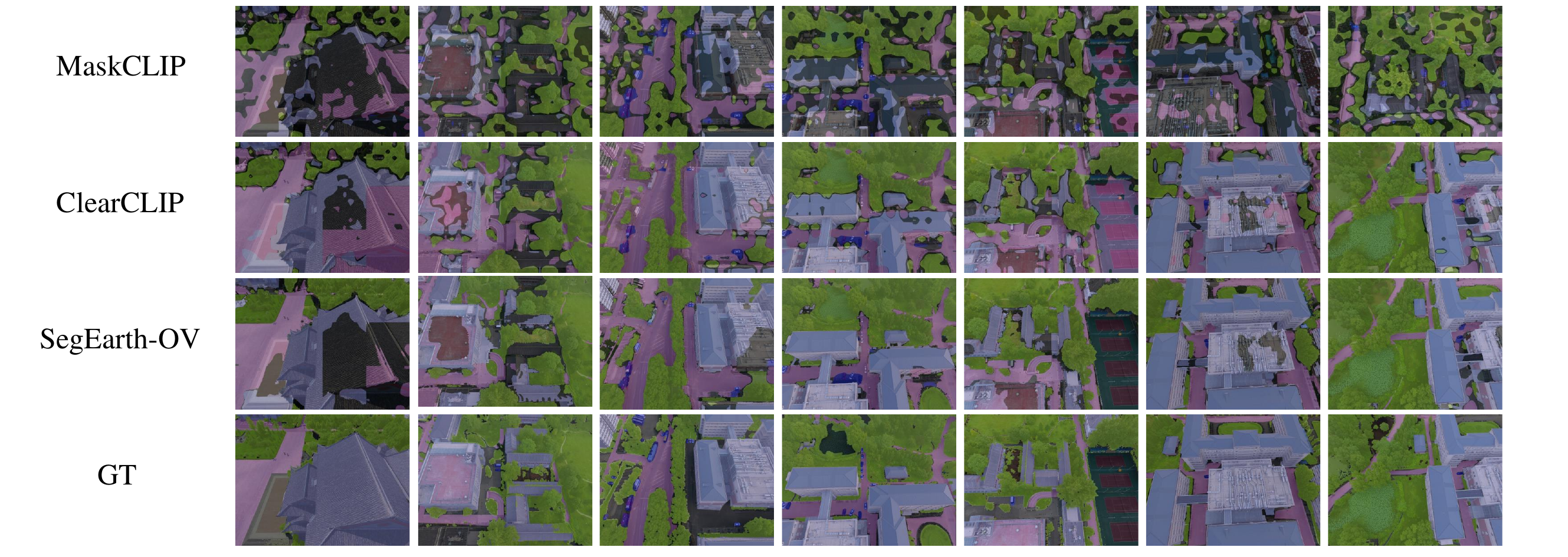}
   \caption{Qualitative comparison between different training-free OVSS methods on UDD5.}
   \label{fig:UDD5}
\end{figure*}

% \begin{figure*}[t]
%   \centering
% %   \fbox{\rule{0pt}{2in} \rule{0.9\linewidth}{0pt}}
%    \includegraphics[width=1.0\linewidth]{figures/Potsdam.pdf}
%    \caption{}
%    \label{fig:Potsdam}
% \end{figure*}

\begin{figure*}[t]
  \centering
%   \fbox{\rule{0pt}{2in} \rule{0.9\linewidth}{0pt}}
   \includegraphics[width=0.8\linewidth]{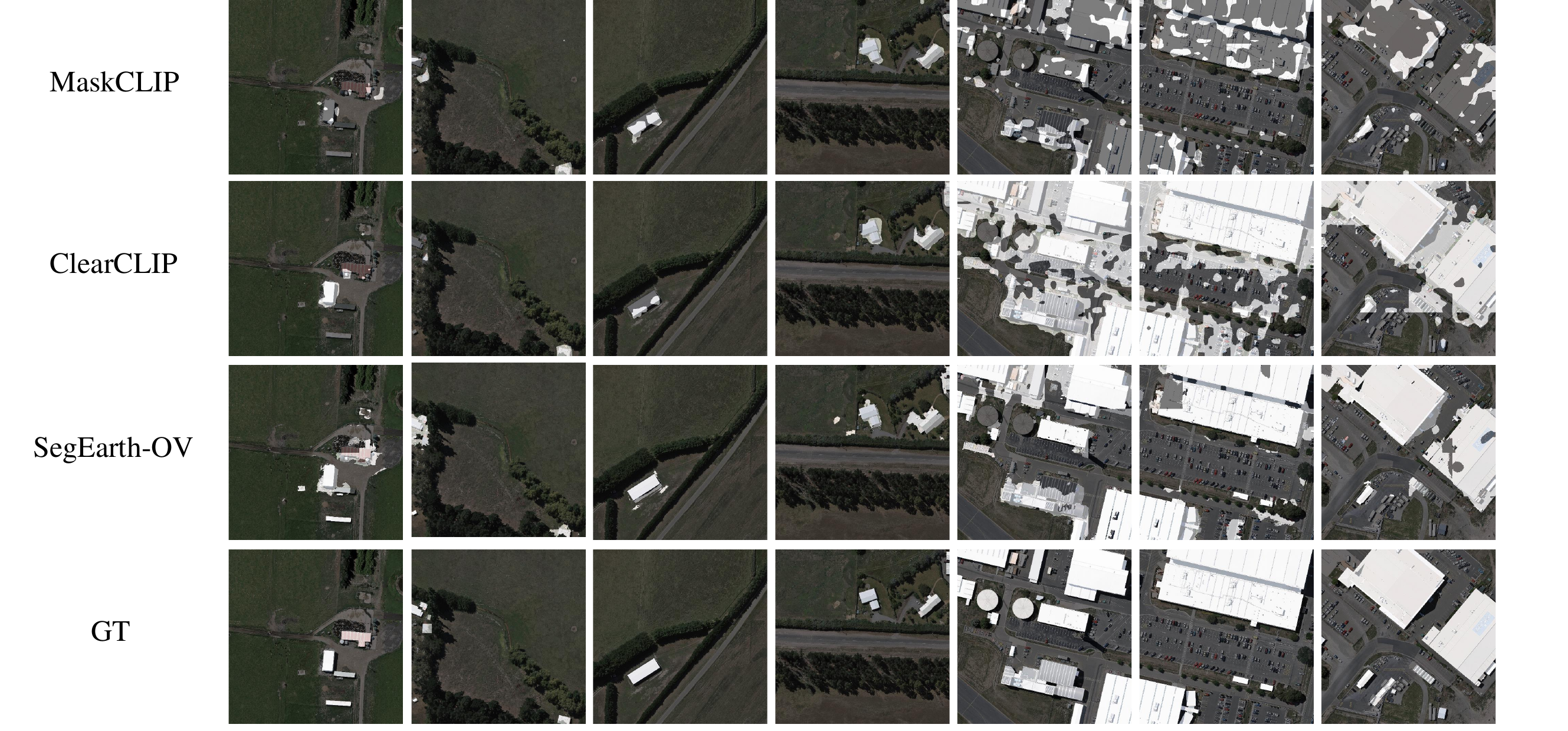}
   \caption{Qualitative comparison between different training-free OVSS methods on WHU$^{Aerial}$.}
   \label{fig:WHU}
\end{figure*}

% \section{Remote sensing CLIP for OVSS}
% \label{sec:rs_clip}

% Some works use remote sensing data for language-image pre-training and yield several remote sensing CLIPs. Typical works include RemoteCLIP \cite{liu2024remoteclip}, GeoRSCLIP \cite{zhang2023rs5m}, and SkyCLIP \cite{wang2024skyscript}, which are trained on 0.8M, 5M, and 5.2M remote sensing data, respectively, without changing the model structure of CLIP. Exploring the capabilities of these remote sensing CLIPs for OVSS on remote sensing images is meaningful. Since these works do not provide the ViT-B/16 version of the image encoder, which is commonly used in OVSS, we uniformly use ViT-B/32. Therefore, we repeat the JBU operation $5\times$ in ``JBU\_One''. For fair comparison, we train the respective SimFeatUp for each model. As listed in \cref{table_rsclip}, RemoteCLIP performs suboptimally to vanilla CLIP, which indicates that a small amount of domain data diminishes the model's transfer capability. GeoRSCLIP achieves the best performance against SkyCLIP, one possible reason is that the images and descriptions used in GeoRSCLIP are more diverse. From another perspective, the OVSS task can be a criterion to assess remote sensing VLMs.

% \section{Natural Images}
% \label{sec:natural}

\end{document}